\pdfoutput=1

\documentclass[11pt]{article}

\usepackage{authblk}
\usepackage[final]{EMNLP2023}

\usepackage{times}
\usepackage{latexsym}

\usepackage[T1]{fontenc}

\usepackage[utf8]{inputenc}

\usepackage{microtype}
\usepackage{multirow}
\usepackage{graphicx}
\usepackage{bm}
\usepackage{amsmath}
\usepackage{amssymb}

\usepackage{multicol}
\usepackage{caption}
\usepackage{subcaption}

\title{P5: Plug-and-Play Persona Prompting for Personalized Response Selection}

\author[1]{Joosung Lee}
\affil[1]{NAVER} 
\author[ ]{Minsik Oh}
\author[2]{Donghun Lee}
\affil[2]{Kakao Brain}
\affil[ ]{\normalsize \texttt{rung.joo@navercorp.com}}

\begin{document}
\maketitle
\begin{abstract}
The use of persona-grounded retrieval-based chatbots is crucial for personalized conversations, but there are several challenges that need to be addressed. 1) In general, collecting persona-grounded corpus is very expensive. 2) The chatbot system does not always respond in consideration of persona at real applications. To address these challenges, we propose a plug-and-play persona prompting method. Our system can function as a standard open-domain chatbot if persona information is not available. We demonstrate that this approach performs well in the zero-shot setting, which reduces the dependence on persona-ground training data. This makes it easier to expand the system to other languages without the need to build a persona-grounded corpus. Additionally, our model can be fine-tuned for even better performance. In our experiments, the zero-shot model improved the standard model by 7.71 and 1.04 points in the original persona and revised persona, respectively. The fine-tuned model improved the previous state-of-the-art system by 1.95 and 3.39 points in the original persona and revised persona, respectively. To the best of our knowledge, this is the first attempt to solve the problem of personalized response selection using prompt sequences.  Our code is available on github~\footnote{https://github.com/rungjoo/plug-and-play-prompt-persona}.
\end{abstract}

\section{Introduction}
Designing a system that naturally communicates with humans is of great interest to researchers and is widely applied to services such as Apple Siri and Amazon Alexa. One of the critical algorithms of these services is multi-turn response selection, which selects the most appropriate response among many response candidates. Selecting a personalized response with a customized chatbot is necessary for a more human-like conversational system. Indeed, \citet{zhangetal2018personalizing} shows that dialog context alone is insufficient for response selection.

\citet{zhangetal2018personalizing} released PERSONA-CHAT where the speakers have each persona. The persona is expressed in multiple sentences, and they get to know each other through conversation. PERSONA-CHAT can be used for research on personalized response generation and selection. However, the following challenges exist in developing personalized response selection for a real application. 1) Building conversations based on Persona is very expensive. PERSONA-CHAT is data from the research environment in English, and persona-grounded corpus in other languages is difficult to access. That is, there is a challenge to build data for real applications. 2) In general domains, the persona may not need to be reflected. However, since previous approaches~\cite{DIM, RSMDCK, CSN, FIRE, BERTCRA, COSPLAY, dasetal2022using} are trained in combination with personas, the model can select a response only given a persona. Therefore, previous approaches always have the disadvantage of reflecting personal information. For example, when the persona is related to a favorite food, it is not helpful knowledge when answering the other topics (i.e., weather). Ideally, a chatbot system needs the ability to consider persona as an option while maintaining standard response selection capabilities.

We propose P5 (\textbf{P}lug-and-\textbf{P}lay \textbf{P}ersona \textbf{P}rompting for \textbf{P}ersonalized Response Selection) to solve the above challenges. First, we assume that there is no expensive persona-based corpus. Therefore, we can train only standard response selection models that do not consider persona. Then, we show that the standard response selection model combined with persona prompting allows response selection to reflect persona, which is a zero-shot inference strategy. Persona prompting improves the performance of standard response selection in persona-based conversations. Also, the model uses persona prompting as optional information because it is a plug-and-play method. If no persona is given to the model, the model acts as a standard response selection model. So we can optionally combine model and persona. Persona sentences to be used for prompting are selected by measuring the similarity to the response. We use a pre-trained model as our similarity model. Only top-$k$ persona sentences are used in order of highest similarity score. In addition, we introduce a zero-shot baseline SoP (Similarity of Persona) based on the similarity score.

To our best knowledge, previous studies only provide fine-tuned models. For comparison in these same experimental settings, we show the experimental results for fine-tuned P5. Our method further improves the performance of the fine-tuned strategy as well as the zero-shot strategy. Fine-tuned P5 achieves state-of-the-art, which proves that persona prompting is effective in learning the relationship between persona and context. We evaluate our methods on PERSONA-CHAT~\cite{zhangetal2018personalizing} and Focus~\cite{focus}. PERSONA-CHAT provides 19 negative responses and 1 positive response for personalized response selection. Focus is given only one positive response as response candidates. Therefore, we build the data by sampling 19 negative candidates.

\section{Related Work}

\subsection{Standard Response Selection}
In dialog systems, retrieval-based response selection is an important module. Earlier retrieval-based methods~\cite{single1, single2} attempted to select a response based on a single-turn context. However, since multi-turn response selection is a scenario for a more realistic service, recent studies~\cite{sabert, UMS, bertfp} focus on selecting a response based on multi-turn context. These multi-turn response selection models leverage pre-trained language models to reflect context. It also improves performance by training a language model to understand conversations through a post-training strategy or multi-task learning. These studies are generally conducted on ubuntu~\cite{ubuntu}, douban~\cite{douban}, and e-commerce~\cite{ecommerce} corpus. Since these datasets are not given a persona, we refer to relevant studies as standard response selection models.

\subsection{Personalized Response Selection}
\label{sec:PMRS}
Standard response selection models suffer from a coherent personality by being trained on general conversations. \citet{zhangetal2018personalizing} releases PERSONA-CHAT dataset, which is a dialogue corpus built on persona. \citet{focus} releases Focus dataset, which is a dialogue corpus built on persona and knowledge.

Recently, many studies introduce fine-tuned models in PERSONA-CHAT:


\citet{RSMDCK} proposes an approach that detects only context-related persona. \textbf{RSM-DCK} (Response Selection Model that can Detect the relevant parts of the Context and Knowledge collection) introduces context selectors and knowledge selectors, which are soft-selection of persona through attention weights. \citet{FIRE} also performs soft-selection of persona, and iteratively referring not only between context and response representations but also between knowledge and response representations to collect deep matching features for scoring response candidates (\textbf{FIRE}: Filtering before Iteratively REferring). \citet{CSN} introduces hard-selection of context-related persona, and shows that recent utterances in context are more important for response selection (\textbf{CSN}: Content Selection Network). \citet{BERTCRA} shows that partner-persona as well as self-persona are important for response selection. \textbf{BERT-CRA} (BERT with Context-Response-Aware Persona) also achieves high performance by combining persona with BERT for context-aware-persona. \citet{COSPLAY} suggests \textbf{COSPLAY} (COncept Set guided PersonaLized dialogue generation Across both partY personas), which considers both speakers as "team". COSPLAY utilizes both self-persona and partner-persona, and proposes a Concept Set framework with a suite of knowledge-enhanced operations to process them such as set algebras, set expansion, and set distance. \citet{dasetal2022using} first learns emotion and intent classifiers respectively with external data. Then, \textbf{BERT-EmA} (Emotion Aware Fusion) and \textbf{BERT-P-EnA} (Entailment Aware Fusion) are learned by adding the predicted emotion and intent of the utterance as input of BERT. These methods are based on BERT-CRA.

\begin{figure*}
    \centering
    \begin{subfigure}{.28\textwidth}
      \centering
      \includegraphics[width=0.9\linewidth]{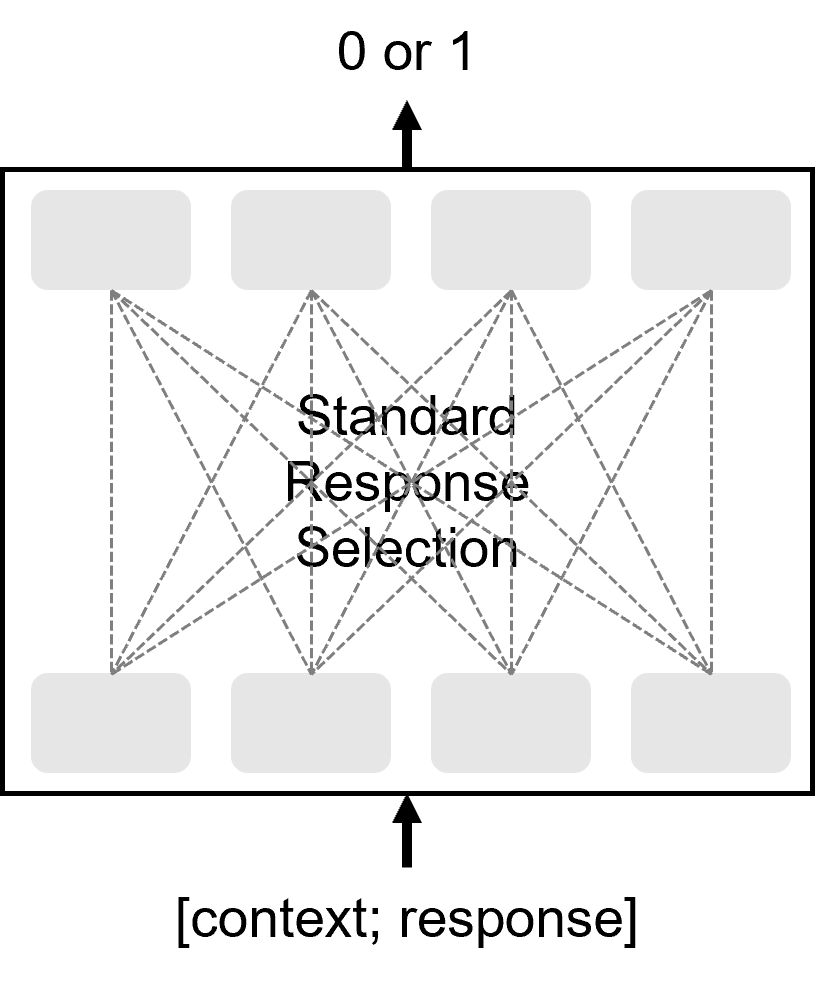}
      \caption{Training phase}
      \label{fig:sfig1}
    \end{subfigure}%
    \begin{subfigure}{.72\textwidth}
      \centering
      \includegraphics[width=0.9\linewidth]{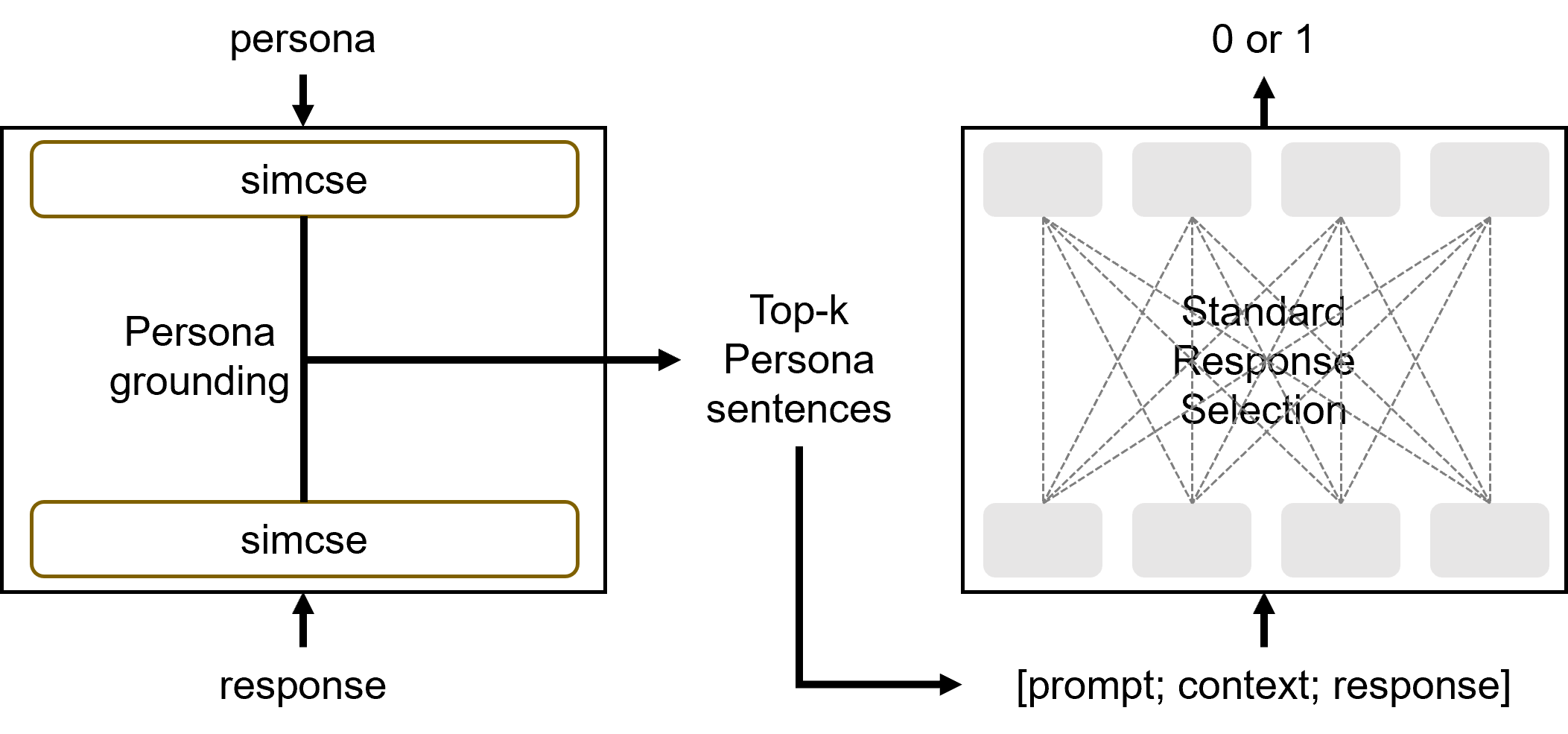}
      \caption{Test phase}
      \label{fig:sfig2}
    \end{subfigure}
\caption{The overview architecture of our proposed P5 model}
\label{fig:overview}
\end{figure*}

\section{Task Definition}
In the training phase, the dialogue dataset $D_{train} = \{c, r, y \}$ is given where context $c = \{u_1, u_2, \cdots, u_n \}$, $u_i$ is the $i$th utterance, $r$ is a response candidate, and $y \in \{0, 1\}$ is a label. $y$ = 1 indicates a positive response, the opposite is $y$ = 0. In the test phase, the dialogue dataset $D_{test} = \{c, r, p, y \}$ is given, where persona $p = \{p_1, p_2, \cdots, p_m \}$ are $m$ persona sentences. Our goal is to learn a matching model without persona from $D_{train}$ and select an appropriate response by reflecting persona from $D_{test}$.

\section{Approach}
Figure~\ref{fig:overview} shows the proposed approach. In the training phase, the model is trained to perform a multi-turn response selection task with $D_{train}$ similar to \citet{sabert, bertfp}, which is called standard response selection. Since our goal is not to improve the performance of standard response selection, we do not use any particular strategy (i.e., post-training). The test phase consists of two steps. The first step is persona grounding corresponding to the response. The second step is calculating the matching score of $(c, r, p)$ with the persona-prompted standard response selection model. Our approach improves the performance by combining the persona prompting with the standard response selection not trained with persona, which can also be utilized as a fine-tuned method.

\subsection{Standard Response Selection (SRS)}
The standard response selection model is trained with $D_{train}$ without a persona. The standard response selection model follows pre-trained language models such as BERT~\cite{bert} and RoBERTa~\cite{roberta}, which are Transformer encoders~\cite{NIPS2017_3f5ee243}. We use RoBERTa as the backbone and train with binary classification for multi-turn response selection. The input format ($x_{st}$) is as follows:


\begin{multline}
x_{st} = \text{concat} ([SEP], u_1, [SEP], u_2, \cdots, \\ [CLS], [SEP], r)
\label{eq:standard_input}
\end{multline}
where $u_i$ is $i$th utterance, [SEP] is a special token to distinguish each utterance, and [CLS] is prepended before the response. Most studies using pre-trained language models for response selection tasks prepend [CLS] to the front of the entire input sequence. However, since the distance between [CLS] and the response changes dynamically, an additional special token must be used to inform which response to classify. We intuitively change the position of [CLS] to represent the input as Equation~\ref{eq:standard_input}.

The response score using the standard response selection model is calculated as follows:
\begin{equation}
s_{st} = W(\text{PLM} (x_{st}))
\label{eq:standard_output}
\end{equation}
where PLM is a pre-trained language model used in the standard response selection model, and $W$ is a matrix that projects the output vector of [CLS] into a two-dimensional vector. $s_{st}$ is the score vector corresponding to $x_{st}$. 

The standard response selection model is trained to minimize cross-entropy loss:
\begin{equation}
L = \frac {1} {N} \sum^{N}_{j=1} \text{CE} (s_{st}^{j}, y^{j})
\label{eq:loss}
\end{equation}
where $j$ means the $j$th training sample, and $N$ is the number of training data.

\begin{table}[!t]
\centering
\resizebox{1.0\columnwidth}{!}{
\begin{tabular}{|c|ccc|cc|}
\hline
\multirow{2}{*}{the number of (\#)}     & \multicolumn{3}{c|}{PERSON-CHAT}                              & \multicolumn{2}{c|}{Focus}        \\ \cline{2-6} 
                                & \multicolumn{1}{c|}{Train} & \multicolumn{1}{c|}{Val}  & Test & \multicolumn{1}{c|}{Train} & Val  \\ \hline
\# conversations & \multicolumn{1}{c|}{8,939}  & \multicolumn{1}{c|}{1,000} & 968  & \multicolumn{1}{c|}{12,484} & 1,000 \\ \hline
\# turns         & \multicolumn{1}{c|}{65,719} & \multicolumn{1}{c|}{7,801} & 7,512 & \multicolumn{1}{c|}{70,332} & 5,639 \\ \hline
\end{tabular}
}
\caption{Statistics of the two datasets}
\label{Tab:dataset}
\end{table}


\subsection{Persona Grounding}
\label{sec:grounding_persona}
\citet{DGMN, DIM, RSMDCK, BERTCRA} use all given persona sentences as an input regardless of context and response. Instead, these approaches perform soft-selection of a persona by assigning attention weights between persona sentences and context embeddings. In other words, small weights are assigned to less relevant persona sentences, affecting response selection less.

\citet{FIRE, CSN} mention that assigning a low weight to a less relevant persona is possible, but the cumulative weight of an irrelevant persona can be significant. Therefore, persona sentences used through the threshold are hard-selected based on the attention weight, which is an essential key for the model. However, since the hard-selection method extracts feature vectors through attention between persona and context, persona sentences are essential inputs for the trained model. That is, the previous frameworks require at least one persona sentence, which is different from selectively combining persona sentences like in our framework. Also, the hard-selection method requires training data, and our persona grounding is sufficient without training data.

We introduce an approach to select only the top-$k$ persona sentences simply and efficiently before combining the standard response selection model and persona in the test phase. A personalized response contains relevant personal information. Therefore, we find the persona sentence related to the response through the similarity between the response and the persona sentence.

\begin{align}
e_r &= \text{SimModel} ([CLS], r) \\
e_{p_i} &= \text{SimModel} ([CLS], p_i) \\
s_{rp_{i}} &= \frac {e_r \cdot e_{p_i}}{||e_r|| \cdot ||e_{p_i}||}
\label{eq:embedding_rp}
\end{align}
where $r$ is the response, $p_i$ is the $i$th persona sentence, $e_r$ and $e_{p_i}$ are the output vectors of [CLS] passed through the similarity model, and $s_{rp_{i}}$ is the similarity score between the two vectors.

A persona sentence with a high similarity score ($s_{rp_{i}}$) is considered to help select the corresponding response. Therefore, we combine only the top-$k$ persona sentences with high similarity scores with the standard response selection model. In our default setting, $k$ = 2, but it can be set dynamically. We used unsupervised simcse~\cite{simcse} and supervised bert-nli~\cite{bertnli} as models for calculating similarity, and the difference between the two is introduced in Section~\ref{sec:effects_persona_grounding}.

\subsection{Persona Prompting}
Recently, GPT-3~\cite{GPT3} model leverages the natural-language prompt to improve few-shot performance very effectively. \citet{gaoetal2021making} achieves high performance by prompt-based fine-tuning of small language models on a small number of training data, which is a more practical scenario. \citet{hanetal2022meet} introduces a dialog prompt that is created using several utterances of fictional characters, in which a pre-trained language model that is not trained in character styles generates attractive responses that mimic the characters.

Inspired by these studies, we propose a persona prompting for personalized response selection. The prompt sequence asks and answers the speaker's persona, simply composed of a prompt question and persona sentences. So the input format ($x_p$) is as follows:

\begin{multline}
x_p = \text{concat} ([SEP], p_q, [SEP], p_1, p_2, \cdots, \\ [SEP], u_1, [SEP], u_2, \cdots, [CLS], [SEP], r)
\label{eq:persona_input}
\end{multline}
where $p_q$ is a prompt question,  which defaults to \textit{"what is your personality?"} sentences are used and $p_i$ ($i \in \{1,...,k\}$) are grounded persona sentences that result from Section~\ref{sec:grounding_persona}. Other prompt questions are described in Section~\ref{sec:other_prompt}.

In Equation~\ref{eq:standard_output}, the input is changed to $x_p$, and the response score is calculated as:

\begin{equation}
s_{p} = W(\text{PLM} (x_p))
\label{eq:persona_output}
\end{equation}
where $s_p$ is the score vector of the response considering persona and context. The response selection model does not learn about persona fusion but naturally recognizes the prompt sequence as part of the context.

\subsection{Baseline: Similarity of Persona (SoP)}
\label{sec:baseline}
In the previous approaches, only fine-tuned approaches have been studied under the assumption that persona is given. Therefore, we introduce a baseline for a simple zero-shot setting, which utilizes a similarity score used to find a persona related to a response. That is, the final response score is the weighted sum of the response score of the standard response selection model and the similarity scores.

\begin{equation}
s_{f} = s_{st} + \alpha F(s_{rp_1}, \cdots, s_{rp_m})
\label{eq:sop}
\end{equation}
where $s_{rp_i}$ is obtained from Equation~\ref{eq:embedding_rp} as a similarity score between the $i$th persona sentence and the response, and $s_{st}$ is obtained from Equation~\ref{eq:standard_output} as a score using the standard response selection model, and $s_f$ is the final score. $F$ is the function to aggregate $s_{rp_i}$ and $\alpha$ is the weight. We tried top-$k$ average function for $F$, but it was most effective to use top-1 $s_{rp_i}$. Therefore, the max function is used for $F$ in the experiment.

\section{Experiments}
\subsection{Datasets}
We experiment on two benchmark datasets. Table~\ref{Tab:dataset} shows the statistics of the datasets, and Table~\ref{Tab:personchat_ex},~\ref{Tab:focus_ex} in Appendix~\ref{sec:dataset} are examples.

\textbf{PERSONA-CHAT} The first dataset is PERSONA-CHAT, where each speaker is described with multiple persona sentences. PERSONA-CHAT is a dataset mainly used in previous studies, and 1 positive response and 19 negative response candidates corresponding to the context. Response selection is performed for every turn of dialogue. PERSONA-CHAT provides two versions of persona, original and revised. The revised persona is data that makes the task more difficult by rephrasing the original persona.

\textbf{Focus} The second dataset is Focus, a dialogue created using persona and knowledge. Focus was created from the motivation that more appropriate utterances can be generated by considering persona and knowledge together. However, for personalized response selection, we only use persona. Since only a positive response is given to the context in Focus, 19 negative response candidates are sampled and formatted according to the response selection task. The sampling strategy follows two steps. (1) Context sampling from the speaker's previous utterances. In this case, utterances using the same persona sentence are sampled first. (2) Random sampling from the corpus. In (1), 2 candidates are sampled, and in (2), 17 candidates are sampled. Therefore, models can achieve high performance by considering both the appropriateness of response and persona in Focus. Also, Focus has a label for persona grounding when constructing a positive response, so it is used to measure the performance of our proposed persona grounding. There can be multiple persona sentences labeled as "True".

\begin{table*}[!t]
\centering
\resizebox{1.7\columnwidth}{!}{
\begin{tabular}{c|c|c|c|c}
\hline
Method                     & Model                           & Original Persona              & Revised Persona               & Focus               \\ \hline\hline
                             & RSM-DCK~\cite{RSMDCK} & 79.65 & 71.85 &                     \\ 
                             & FIRE~\cite{FIRE}                            & 81.6                          & 74.8                          &                     \\                              
                             & CSN-word~\cite{CSN}                        & 78.1                          & 70.1                          &                     \\ 
                             & BERT-CRA~\cite{BERTCRA}                        & 84.3                          & 79.4                          &                     \\ 
                             & COSPLAY~\cite{COSPLAY}                         & 85.5                          & 74.4                          &                     \\ 
                             & BERT-EmA~\cite{dasetal2022using}                         & 84.6                          & 79.8                          &                     \\ 
                             & BERT-P-EnA~\cite{dasetal2022using}                         & 85.3                          & 80.5                          &                     \\ 
\multirow{-8}{*}{Fine-tuning} & \textbf{P5}                     & \textbf{87.45}                & \textbf{82.79}                & \multirow{-7}{*}{-} \\ \hline\hline
                             & SRS                           & 72.4                         & 72.4                         & 91.65               \\   
                             & SoP                             & 76.41                         & 73.16                         & 94.27               \\ 
\multirow{-3}{*}{Zero-shot}  & \textbf{P5}                     & \textbf{80.11}                & \textbf{73.44}                & \textbf{97.85}      \\ \hline
\end{tabular}
}
\caption{Evaluation results on the test sets of PERSONA-CHAT and validation sets of Focus. Performance is measured as $R@1$. Bold text indicates the best performance in each part. In PERSONA-CHAT and Focus, RoBERTa-base are used as PLM.}
\label{Tab:results}
\end{table*}

\subsection{Evaluation Metric}
We use the evaluation metric used in previous works~\cite{BERTCRA, FIRE} for a fair comparison. Each model checks whether the candidate with the highest ranking score is a positive response, denoted by $R@1$. Specifically, both PERSONA-CHAT and Focus are $R_{20}@1$ because 1 positive response and 19 negative responses are given as candidates.

\subsection{Training Setup}
We use a pre-trained model from the huggingface library~\footnote{https://huggingface.co/docs/transformers/index}. For the standard response selection model, we use AdamW as the optimizer. The learning rate is an initial value of 1e-6, and \text{get\_linear\_schedule\_with\_warmup} provided by the huggingface library is used for the learning rate scheduler. The maximum value of 10 is used for the gradient clipping.  The training epoch is 10, the model is evaluated on the validation data for each epoch, and the best model is selected. In prompt-based fine-tuning, the training epoch is 5, and the rest are the same as the standard response selection. All experiments are performed on one A100 GPU, and the results are for a single turn because there is little variation between each run.

\subsection{Results}
Table~\ref{Tab:results} shows the evaluation results of the previous and the proposed method for two benchmarks. Previous methods are described in Section~\ref{sec:PMRS}. SRS is a standard response selection model that does not consider persona. SoP is a baseline using the similarity score of persona introduced in Section~\ref{sec:baseline}. P5 is our proposed method using persona prompting.

SRS does not consider persona but achieves a performance of 72.4 in PERSONA-CHAT, which is an unsatisfactory performance in the original persona, but a good performance in the revised persona. We believe that many previous models (RSM-DCK, FIRE, CSN-word, COSPLAY) were effective at fusing the original persona, but did not fuse the revised persona well.

SoP is a simple baseline we propose, which improves the performance of SRS. In Equation~\ref{eq:sop}, $\alpha$ is 0.5 and 0.05 in the original persona (or Focus) and revised persona, respectively, and an appropriate value for $\alpha$ was selected through an experiment. Since the original persona has many examples that directly overlap the words of the response, the similarity score is more effective. However, since a revised persona is a rephrased sentence, simply scoring similarity with persona sentences is less effective.

P5 identifies the speaker's persona from the prompt in the form of simple dialogue and uses it for response ranking. P5 achieves the best performance in both zero-shot and fine-tuning. Zero-shot P5 improves the performance of the SRS through persona prompting by 7.71 points in the original persona and 1.04 points in the revised persona. The zero-shot inference strategy is more effective for the original persona (or Focus) than the revised persona, which is considered difficult for the SRS to understand the revised persona as a prompt context. When a persona is given as training data, the fine-tuned P5 achieves 87.45 and 82.79 performance in the original persona and revised persona, respectively, which is a remarkable performance improvement compared to previous models. Also, our proposed prompt is a plug-and-play module and has the advantage of being turned on and off according to the real application.

We further experimented with the zero-shot setting in Focus to verify the effectiveness of our model. Focus is structured similarly to the original persona in PERSONA-CHAT, and the experimental results also show the same aspect as the original persona. Since zero-shot P5 has already achieved satisfactory performance in Focus, fine-tuned P5 has not been tested.

\begin{figure}[!t]
    \centering 
    \includegraphics[width=1.0\columnwidth]{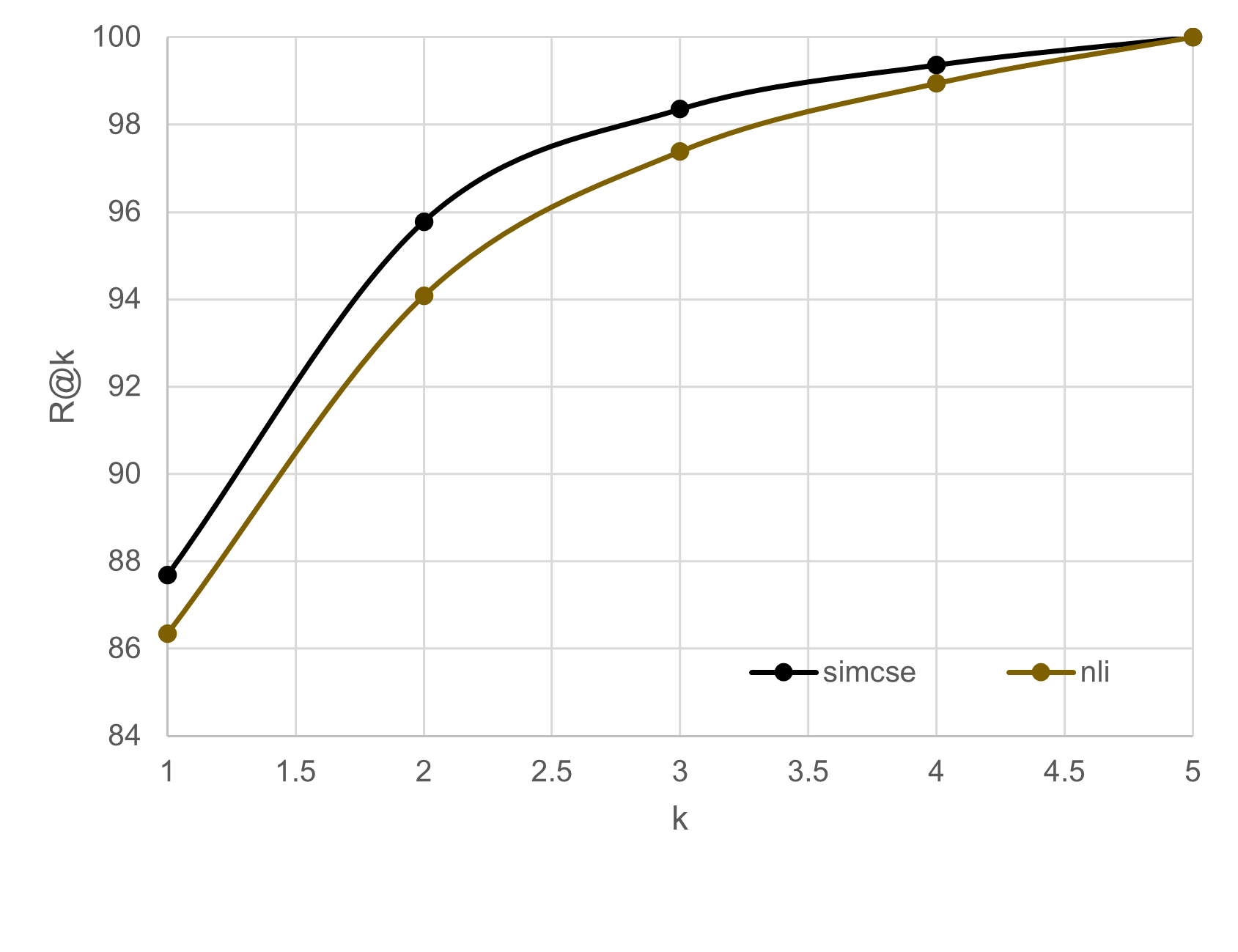}
    \caption{Performance of similarity model on persona grounding in Focus}
    \label{fig:topk}
\end{figure}

\begin{table*}[!htb]
\centering
\resizebox{2.0\columnwidth}{!}{
\begin{tabular}{c|c|c|c|c|c}
\hline
Model               & prompt question              & similarity model & order of persona & Original Persona & Revised Persona \\ \hline\hline
\multirow{7}{*}{Zero-shot P5} & \textbf{\textit{what is your   personality?}}  & \textbf{simcse}           & \textbf{ascending}        & \textbf{80.11}            & \textbf{73.44}           \\ 
                    & \textit{tell me your   personality.}  & simcse           & ascending        & 79.73            & 73.34           \\
                    & \textit{tell me more about yourself.} & simcse           & ascending        & 79.81            & 73.32           \\
                    & random utterance & simcse           & ascending        & 78.89            & 72.63           \\
                    & empty string & simcse           & ascending        & 79.71            & 73.31           \\
                    \cline{2-6} 
                    & \textit{what is your   personality?}  & nli              & ascending        & 79.86            & 73.4           \\ \cline{2-6} 
                    & \textit{what is your   personality?}  & simcse           & descending       & 79.7            & 73.34           \\ \hline
\end{tabular}
}
\caption{Performance with variants of persona prompting}
\label{Tab:prompting}
\end{table*}

\subsection{Effects of Persona Grounding}
\label{sec:effects_persona_grounding}
With a persona format similar to the original persona in PERSONA-CHAT, Focus provides a persona grounding label. Figure~\ref{fig:topk} shows the performance of persona grounding using simcse and bert-nli models in Focus, where both simcse and bert-nli models are pre-trained models. The evaluation metric is $R@k$, where the value is considered 100 if the number of "True" candidates is 0. As $k$ increases, $R@k$ improves, which increases the probability that persona sentences related to the response are reflected. However, as $k$ increases, irrelevant persona sentences are also entered as input. Therefore, an appropriate value of $k$ is required.

\begin{figure}[!t]
    \centering 
    \includegraphics[width=1.0\columnwidth]{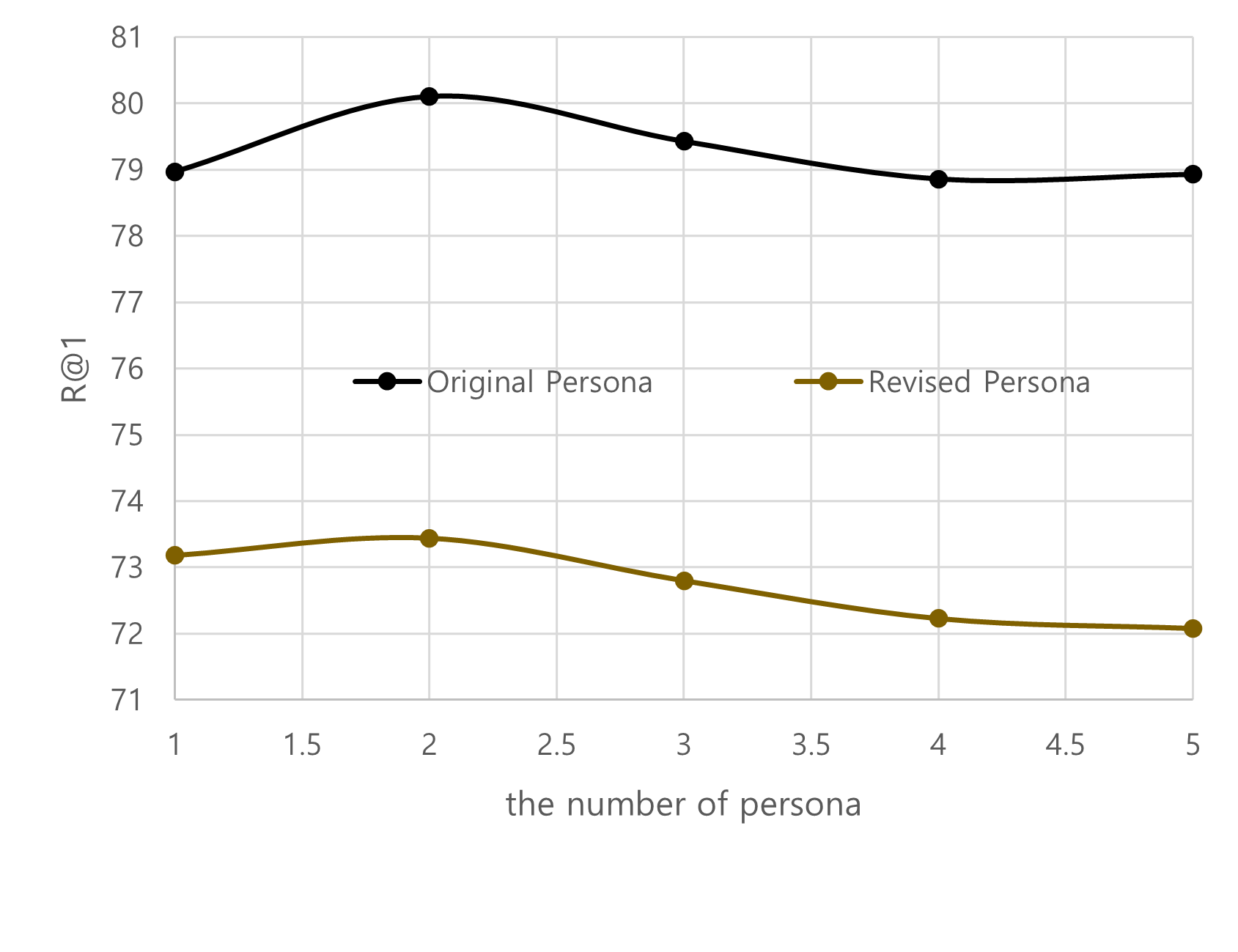}
    \caption{Performance of zero-shot P5 change according to the number of persona sentences used}
    \label{fig:numpersona}
\end{figure}

In Table~\ref{Tab:results}, P5 is the evaluation results using only 2 persona sentences (top-2). Since PERSONA-CHAT does not have a label for persona grounding, we cannot confirm the ground truth persona sentences reflected in the response. Figure~\ref{fig:numpersona} is zero-shot P5 performance change according to the number of persona sentences used in PERSONA-CHAT. The best performance is achieved when two persona sentences are used in both the original persona and the revised persona. More persona sentences increase the persona grounding performance but confuse the personality to be reflected by the model and degrade the performance. Simcse has a higher rate of finding used persona sentences than the bert-nli model, and similar results can be expected in PERSONA-CHAT. Comparing the 1st and 6th rows in Table~\ref{Tab:prompting}, it can be seen that simcse is more effective than bert-nli in PERSONA-CHAT. 


\subsection{Structure of Persona Prompting}
\label{sec:other_prompt}
Table~\ref{Tab:prompting} shows zero-shot P5 performance for the variant of persona prompting. We changed the prompt question \textit{"what is your personality?"} to the rephrase \textit{"tell me your personality."} and \textit{"tell me more about yourself."}. The performance difference according to the prompt question is not large, and it is not easy to find the optimal prompt question in the discrete space.

The random utterance (4th row) indicates that the prompt question was randomly sampled from the training utterances. The empty string (5th row) indicates that there is no prompt question, which means that the prompt sequence consists only of persona sentences. These two methods make it difficult to know whether the persona sentences represent the speaker's personality. Performance is slightly lower than when the prompt question is \textit{"what is your personality?"}, but it doesn't show a huge difference. That is, persona sentences are more important to performance than prompt questions.

We also experiment with two methods for the input sequence of a grounded persona. The first is the ascending method, from lowest to highest similarity score, which means that the position of the most similar persona and response is close (1st row). The second is the descending method, from highest to lowest similarity score, which means that the position of the most similar persona and response is far (7th row). Experimental results show that the ascending method achieves higher performance. Therefore, the closer the distance between the relevant persona and the response, the more effective it is for the model.

\subsection{Other Standard Response Selection Model}
\label{sec:other_SRS}
Table~\ref{Tab:large_result} shows the results for another backbone in PERSONA-CHAT. We experimented by changing the backbone of the SRS from RoBERTa-base to RoBERTa-large. The performance of the SRS-large is better than that of the SRS-base. Zero-shot P5-large improves the performance by 7.17 and 1.65 points, respectively, in the original persona and revised persona compared to SRS. Regardless of the performance of the SRS, the P5 effectively combines persona to improve performance. Table~\ref{Tab:backbone} in Appendix~\ref{sec:backbone} show experiments on more diverse backbones. 

Fine-tuned P5-base achieves the best performance, but fine-tuned P5-large achieves SoTA with a larger margin. In addition, zero-shot P5-large achieves competitive performance with previous fine-tuning approaches. That is, with better SRS, we observe that even the zero-shot approach can achieve remarkable performance.


\begin{table}[!t]
\centering
\resizebox{1.0\columnwidth}{!}{
\begin{tabular}{c|c|c|c}
\hline
Method                     & Model (large)    & Original Persona                                             & Revised Persona                                              \\ \hline\hline
Fine-tuning & P5 & 90.54 & 86.62 \\ \hline\hline
\multirow{2}{*}{Zero-shot} & SRS & 77.96 & 77.96 \\ \cline{2-4} 
                           & P5 & 85.13  & 79.61  \\ \hline
\end{tabular}
}
\caption{Experimental results for a large model.}
\label{Tab:large_result}
\end{table}

\subsection{Ablation Study}
We perform several ablation studies when testing to know the importance of each part of the framework. Without persona grounding (G), P5 considers the personality to be all persona sentences, and the order is randomized to form the prompt sequence. This is similar to the experiment in Figure~\ref{fig:numpersona} where the number of persona = 5. Without prompt question (Q), P5 uses only persona sentences as the prompt sequence, which is the same as the 5th row in Table~\ref{Tab:prompting}. Without a prompt sequence (P), P5 does not consider persona as context, which is equivalent to standard response selection. -D indicates that SRS does not have access to PERSONA-CHAT dialogue, so it is trained with external data. We use dailydialog~\cite{lietal2017dailydialog} as training data and 10 negative candidate responses are randomly sampled. 

In the zero-shot setting, ablation studies are performed on persona grounding, prompt question, prompt sequence (prompt question+persona sentence), and dialogue corpus. All components clearly show differences in the original persona. The absence of persona grounding and prompt questions reduces performance, but these are considered minor components. However, persona sentences are an important component of performance, and using them as prompt sequences is our major contribution. We also assumed that the model is inaccessible to dialogue as well as persona sentences from PERSONA-CHAT. So SRS is trained as dailydialog. Zero-shot P5 (-D\&P) (i.e. SRS w/ dailydailog), without using persona sentences in the test, achieves a performance of 44.5. Zero-shot P5 (-D) utilizing persona sentences achieves the performance of 59.94 and 50.23 in the original persona and revised persona, respectively, which is a much larger performance improvement than shown in Table~\ref{Tab:results}. With the same conclusion as in Section~\ref{sec:other_SRS}, the proposed prompting proves to lead to a large performance improvement regardless of SRS.

In the fine-tuning setting, ablation studies are performed on persona grounding, prompt sequences when testing. Fine-tuning P5 (-G) achieves 87.13 and 81.55 performance on the original and revised, respectively, showing that the performance difference due to persona grounding is smaller than the zero-shot method. In addition, it has the advantage of not requiring additional computation for persona grounding. This is because the model gains the ability to attend to the appropriate persona when selecting a response through learning from the persona corpus. Therefore, our prompting method operates effectively in a fair comparison with other frameworks. Fine-tuning P5 (-P) achieves a performance of 68.18 and 70.38 in the original and revised versions, respectively, which is worse than zero-shot P5 (-P). Therefore, we find that fine-tuning P5 exhibits a strong dependence on the persona sentence when selecting responses. These limitations will have similar limitations to fine-tuned models as in previous studies.

\begin{table}[!t]
\centering
\resizebox{1.0\columnwidth}{!}{
\begin{tabular}{c|c|c|c}
\hline
Method                      & Model      & Original Persona & Revised Persona \\ \hline\hline
                            & P5         & 80.11 & 73.44           \\ \cline{2-4}
                            & -G         & 78.93 & 72.08              \\
                            & -Q        & 79.71 & 73.31            \\
                            & -P        & 72.4 & 72.4            \\ \cline{2-4}
                            & -D         & 59.94 & 50.23            \\
\multirow{-6}{*}{Zero-shot} & -D\&P        & 44.5 & 44.5 \\ \hline\hline
                             & P5 & 87.45 & 82.79               \\ \cline{2-4}  
                             & -G & 87.13 & 81.55               \\ 
\multirow{-3}{*}{Fine-tuning}  & -P & 68.18 & 70.38      \\ \hline

\hline
\end{tabular}
}
\caption{Ablation study for each module of the framework. G stands for persona grounding, Q stands for prompt question, P stands for prompt sequence, and D stands for PERSONA-CHAT dialogue.}
\label{Tab:ablation}
\end{table}

\section{Conclusion}
In this paper, we present a method called P5, which functions as a plug-and-play system that only incorporates persona when desired. Our approach involves identifying related persona sentences through their similarity to a given response, and then adding these sentences as a prompt to the input. This allows the standard response selection model to better match context and response by taking into account the persona. We evaluate our method on two benchmark datasets using both fine-tuning and zero-shot settings. Fine-tuned P5 outperforms previous studies by a significant margin. Zero-shot P5 also effectively improves performance when compared to standard response selection models. Even the zero-shot P5-large shows performance that is comparable to previous fine-tuning approaches.

P5 is only evaluated using persona-based corpus, however, in real-world applications, persona information is not always available. Therefore, it is important that the standard response selection model can be combined with persona in a dynamic manner. One way to achieve this is by only incorporating persona sentences that have a similarity score above a certain threshold. 
We plan to investigate other options for reflecting persona in future studies.

\section*{Limitations}
P5 is only evaluated using persona-based corpus, however, in real-world applications, persona information is not always available. Therefore, it is important that the standard response selection model can be combined with persona in a dynamic manner. One way to achieve this is by only incorporating persona sentences that have a similarity score above a certain threshold. We plan to investigate other options for reflecting persona in future studies.

The importance of a standard response selection model outweighs the use of persona sentences in personalized response selection. In Table~\ref{Tab:ablation}, the P5 (-D) performance improves with persona prompting, however, it is still lower than that of P5 (-P). The low performance of the standard response selection model (P5 (-D\&P)) is the reason for this. To improve zero-shot P5 performance, it is crucial to improve the standard response selection performance. Therefore, we will conduct further research on enhancing the performance of zero-shot standard response selection models that do not utilize PERSONA-CHAT.

\bibliography{anthology,custom}
\bibliographystyle{acl_natbib}

\appendix

\section{Dataset Example}
\label{sec:dataset}

PERSONA-CHAT has \{8,939, 1,000, 968\} conversations in \{train, validation, test\} and Focus has \{12,484, 1,000\} conversations in \{train, validation\}. Table~\ref{Tab:personchat_ex},~\ref{Tab:focus_ex} show a example of PERSONA-CHAT and Focus, respectively.


\begin{table*}[!h]
\centering
\resizebox{1.8\columnwidth}{!}{
\begin{tabular}{c|ll}
\hline
\multicolumn{1}{l|}{}     & \multicolumn{1}{c|}{Pesona 1}                                      & \multicolumn{1}{c}{Persona 2}          \\ \hline
\multirow{5}{*}{Original} & \multicolumn{1}{l|}{i love to meet new   people.}                  & i work as an   accountant.            \\
                          & \multicolumn{1}{l|}{my favorite   sport is ultimate frisbee.}      & i live in ohio.                       \\
                          & \multicolumn{1}{l|}{autumn is my   favorite season.}               & i am a single mom of two boys.        \\
                          & \multicolumn{1}{l|}{my parents   are living in bora bora.}         & i drive a honda civic.                \\
                          & \multicolumn{1}{l|}{i have a   turtle named timothy.}              & i like to go hiking in my spare time. \\ \hline
\multirow{5}{*}{Revised}  & \multicolumn{1}{l|}{i like getting   friends.}                     & they call me a bean   counter.        \\
                          & \multicolumn{1}{l|}{i love to run   around and get out my energy.} & i am from the north.                  \\
                          & \multicolumn{1}{l|}{i love   watching the leaves change colors.}   & i am raising sons all on my own.      \\
                          & \multicolumn{1}{l|}{my family   lives on a island.}                & i own a small car.                    \\
                          & \multicolumn{1}{l|}{reptiles make   good pets.}                    & i enjoy nature walks.                 \\ \hline
\multirow{6}{*}{Dialogue} & \multicolumn{2}{l}{person 1: hi , i am kera   and i am a social butterfly}                                 \\
                          & \multicolumn{2}{l}{person   2: hi . i am more the mousy type . numbers are my world at my day job . you ?} \\
                          & \multicolumn{2}{l}{person   1: i work for a tech firm , i am a tom girl}                                   \\
                          & \multicolumn{2}{l}{person   2: i am just an ohio mom with two amazing sons . not married though .}         \\
                          & \multicolumn{2}{l}{person   1: cool . i have no kids just my pet turtle timothy}                           \\
                          & \multicolumn{2}{l}{person   2: great pet name . i do not have any pets unless you count my car , sally .}  \\ \hline
\end{tabular}
}
\caption{An example from PERSONA-CHAT dataset}
\label{Tab:personchat_ex}
\end{table*}

\begin{table*}[!h]
\centering
\resizebox{1.4\columnwidth}{!}{
\begin{tabular}{c|lc}
\hline
\multicolumn{1}{l|}{}     & \multicolumn{1}{c|}{Persona 2}                                                                        & Persona   Grouding                                                   \\ \hline
\multirow{5}{*}{Original} & \multicolumn{1}{l|}{I like to go to   Church.}                                                        & FALSE                                                                \\
                          & \multicolumn{1}{l|}{I am Roman Catholic.}                                                             & FALSE                                                                \\
                          & \multicolumn{1}{l|}{I wish to go to El Paso.}                                                         & TRUE                                                                 \\
                          & \multicolumn{1}{l|}{I would like to go to Texas.}                                                     & FALSE                                                                \\
                          & \multicolumn{1}{l|}{I love the United States.}                                                        & FALSE                                                                \\ \hline
\multirow{7}{*}{Dialogue} & \multicolumn{2}{l}{person 1: Wow, this is amazing! What is this?}                                                                                                            \\
                          & \multicolumn{2}{l}{person 2: It is a Church, something which you   like.}                                                                                                    \\
                          & \multicolumn{2}{l}{person 1: What is the name of this place?}                                                                                                                \\
                          & \multicolumn{2}{l}{\begin{tabular}[c]{@{}l@{}}person 2: The name of this place is The Roman   Catholic Diocese \\ of El Paso, remember you are Roman Catholic.\end{tabular}} \\
                          & \multicolumn{2}{l}{person 1: Where is this place?}                                                                                                                           \\
                          & \multicolumn{2}{l}{person 2: It is located in El Paso, a city   which you wish to go.} \\ \hline
\end{tabular}
}
\caption{An example from Focus dataset. The label of persona grounding is for the last utterance.}
\label{Tab:focus_ex}
\end{table*}

\section{Additional Standard Response Selection Model}
\label{sec:backbone}

\begin{table*}[!t]
\centering
\resizebox{1.5\columnwidth}{!}{
\begin{tabular}{c|c|c|c|c}
\hline
{ Backbone}                                                & { Method}                                                & { Model}                                        & { Original Persona} & { Revised Persona} \\ \hline\hline
{ }                                & { }                              & { SRS}                                          & { 72.4}            & { 72.4}           \\
 & { }                              & { P5}     & { 80.11}            & { 73.44}           \\ \cline{3-5}
{ }                                & { }                              & { SRS w/ DD}                                          & { 44.5}            & { 44.5}           \\
\multirow{-4}{*}{{ RoBERTa-base}} & { }                              & { P5 w/ DD}                                           & { 59.94}            & { 50.23}           \\ \cline{1-1} \cline{3-5} 
{ }                                & { }                              & { SRS}                                          & { 77.96}            & { 77.96}           \\
\multirow{-2}{*}{{ RoBERTa-large}} & { }                              & { P5}                                           & { 85.13}            & { 79.61}           \\ \cline{1-1} \cline{3-5} 
{ }                                & { }                              & { SRS}                                          & { 69.61}            & { 69.61}           \\
\multirow{-2}{*}{{ BERT-base}}     & { }                              & { P5}                                           & { 76.66}            & { 69.74}           \\ \cline{1-1} \cline{3-5} 
{ }                                & { }                              & { SRS}                                          & { 67.37}             & {67.37}            \\
\multirow{-2}{*}{{ ALBERTv2-base}} & { }                              & { P5}                                           & { 76.58}                & { 68.89}               \\ \cline{1-1} \cline{3-5} 
{ }                                & { }                              & { SRS}                                          & {69.68}            & { 69.68}           \\
\multirow{-2}{*}{{ ConvBERT-base}} &  \multirow{-12}{*}{{ Zero-shot}}                              & { P5}                                           & {78.49}            & { 71.49}           \\ \hline
\end{tabular}
}
\caption{Experimental results from all the backbones we experimented with.}
\label{Tab:backbone}
\end{table*}

We prove that the proposed prompt sequence is effective through more extensive experiments with various backbones. In addition to RoBERTa, we experiment with BERT~\cite{bert}, ALBERT~\cite{albert}, and ConvBERT~\cite{convbert} as backbones. \textbf{ALBERT} (A Lite BERT) is a language model for natural language processing that was designed to be more efficient than its predecessor, BERT. ALBERT reduces the amount of computation through matrix decomposing through factorized embedding parameterization and uses Sentence Order Prediction (SOP) as loss. \textbf{ConvBERT} is a language model for natural language processing that combines the power of convolutional neural networks with transformer architecture. It utilizes a hierarchical convolutional structure to capture local dependencies and long-range dependencies, resulting in improved computational efficiency and better performance on various language tasks. The experimental settings are the same for all backbones.

SRS is a response selection model trained without persona sentences. P5 is a model in which SRS receives both the prompt sequence and context as inputs and selects a response when testing. "with DD" means that the performance was measured based on the SRS learned by dailydialog, which is the same as -D and -D\&P in Table~\ref{Tab:ablation}. Depending on the backbone model, the effect of improving performance is different, but in the original persona, the model's ability improves by a large margin. In the revised persona, P5 performance improves less than in the original persona. The large backbone did not differ from the base backbone. SRS w/ DD trained with dailydialog has lower performance, but the effect of the prompt sequence is significant.

We demonstrate that prompt sequences improve the performance of SRS on all backbones. We believe that pre-trained language models help to perform this zero-shot inference because they are trained from a large corpus and thus have the ability to understand context. The prompt sequence helps to achieve higher performance regardless of the performance of SRS, which is a very simple but powerful method. However, the effect is relatively small in the revised persona, so further research will be done in this area.

\end{document}